\documentclass[a4paper]{article}

\usepackage[english]{babel}
\usepackage[utf8x]{inputenc}
\usepackage[T1]{fontenc}

\usepackage[a4paper,top=3cm,bottom=2cm,left=3cm,right=3cm,marginparwidth=1.75cm]{geometry}

\usepackage{amsmath}
\usepackage{graphicx}
\usepackage{amsthm}
\usepackage{mathrsfs}
\usepackage[colorinlistoftodos]{todonotes}
\usepackage[colorlinks=true, allcolors=blue]{hyperref}

\newtheorem{theorem}{Theorem}
\newtheorem{lemma}{Lemma}
\newtheorem*{NoNumTheorem}{Theorem}

\title{The Force of Proof by Which any Argument Prevails}
\author{Brian Shay and Patrick Brazil\footnote{The main ideas presented here, including axioms for arguments, were developed by the first author of this report  from an imaginative reading of Ars Conjectandi in 2003, after years (1987-2003) of concentrated reading and research in AI topics, principally focused on uncertain reasoning. He is grateful to Rohit Parikh for extensive encouragement throughout this period and dozens of brilliant and amusing lectures.
An attempt to publish these ideas in 2003 failed, and this author walked away from the ``Bernoulli project'' after a disagreement with a referee, who identified no computational errors but failed to see the significance of results,   to concentrate on an  unrelated project.  The author's role in the unrelated project was to provide mathematical and  programming support to the development  by a colleague of a closed form calculus for portfolios of vanilla and exotic options  and underlying securities, dispensing altogether with simulations and numerical methods in pricing and hedging (see e.g. \cite{izmailov2014breakthrough}, \cite{izmailov2015complete}).  Counting on the brilliance of the design and estimating (correctly, it happens) the likelihood of the success of the option project, the author was distracted from completing the ``Bernoulli project'' for over fifteen years, until now.
Substantial  improvements to this ``Bernoulli project'' were added by collaboration of the present authors in connection with the second author's ``case study'' requirement for a Master's Degree in Applied Mathematics in the Department of Mathematics and Statistics,  Hunter College, CUNY in 2017.}
\\
Hunter College CUNY
\\
Department of Mathematics and Statistics
\\
email: bshay@hunter.cuny.edu
\\
email: pbrazil1492@gmail.com}

\begin{document}
\maketitle

\begin{abstract}
Jakob Bernoulli, working in the late 17th century, identified a gap in contemporary probability theory. He cautioned that it was inadequate to specify force of proof (probability of provability) for some kinds of uncertain arguments. After 300 years, this gap remains in present-day probability theory. We present axioms analogous to Kolmogorov axioms for probability, specifying uncertainty that lies in an argument’s inference/implication itself rather than in its premise and conclusion. The axioms focus on arguments spanning two Boolean algebras, but generalizes the obligatory: force of proof of A implies B is the probability of (B or not(A)) in the case that the two Boolean algebras are identical. We propose a categorical framework that relies on generalized probabilities (objects) to express uncertainty in premises, to mix with arguments (morphisms) to express uncertainty embedded directly in inference/implication. There is a direct application to Shafer’s evidence theory (Dempster-Shafer theory), greatly expanding its scope for applications. Therefore, we can offer this framework not only as an optimal solution to a difficult historical puzzle, but to advance the frontiers of contemporary artificial intelligence.
\\
\\
Keywords: force of proof, probability of provability, Ars Conjectandi, non-additive probabilities, evidence theory,  
\end{abstract}

\section*{Introduction}

The title is a nominal description by Jakob Bernoulli of his incomplete and still neglected project, undertaken in the late $17^{th}$ century. From notes preserved in $\textnormal{“Ars Conjectandi”}$ \cite{bernoulli2006art}, his posthumous manuscript (1713), it is clear that Bernoulli thought it useful to assign weights to arguments as an indication of their adequacy, and that these weights should be associated with probabilities of provability.
\vspace{2pt} 

A small number of contemporary mathematicians/statisticians/historians have quarreled over what Bernoulli intended in this discussion (e.g. Hailperin \cite{hailperin1996sentential}, Shafer \cite{shafer2008non}), so there is no clear indication from
Bernoulli himself of a natural path to completing his project.
\vspace{2pt}

We shall not try represent each of Bernoulli’s technical exercises within a new overarching theory, but rather take hints from Bernoulli and his few close critics to formulate a theory of uncertain argumentation in contemporary mathematical terms. It is our hope that this theory will appear to the reader, even the most attentive to historical details, to be consistent with Bernoulli’s aims. Moreover, apart from its role in solving a puzzle of considerable historical interest, our approach to uncertain reasoning specified here will contribute powerful new techniques for explanation-based problem-solving in artificial intelligence, by contrast to “black box” approaches (e.g. machine learning) . Judea Pearl has written a convincing manifesto on this topic in a recent WSJ article aimed at a wide audience \cite{pearl_mackenzie_2018} that ranks explanation-based approaches as inherently higher forms of reasoning, relative to black box approaches whose limits and successes cannot be fathomed. He is echoing concerns of users and investors who are raising questions in business journals about the overheated culture of machine reasoning.
\vspace{2pt}

Hailperin and Shafer focus on different idioms of uncertain reasoning to explain Bernoulli’s ideas about force of proof. Hailperin emphasizes the calculus of expected values, Shafer emphasizes non-additive
probabilities.
\vspace{2pt}

Shafer’s insight seems to be correct, but his technical analysis of Bernoulli’s project is informal and leaves undiscovered a rich formal structure that has eluded Bernoulli’s readers for hundreds of years. We disclose that the (non-normalized) mass functions of Shafer’s evidence theory can play the key role in a formal solution to Bernoulli’s puzzle. These are the canonical non-additive probability models in contemporary reasoning about uncertainty. As an application of our analysis of Bernoulli’s problem, we greatly expand the scope of Shafer’s evidence theory. This expansion is achieved by identifying a “category of belief functions”, within which Shafer’s compatibility relations represent only the most rudimentary morphisms--- and therefore almost all morphisms between belief functions are missing from Shafer’s theory.
\vspace{2pt}

An example of Bernoulli makes it extremely unlikely that Hailperin’s calculus is correct. Bernoulli considers the extent to which his own brother’s laziness accounts for his not writing a letter to him when expected. Bernoulli stipulates that his brother is lazy (assumption A) and that the letter was not written (candidate consequence C). Bernoulli raises the question: what is the force of proof connecting A to C? There is no puzzle if the force of proof is the probability of material implication: probability( C or not(A) ), which is 1. Evidently, this is not the answer to Bernoulli’s question, which lies deeper. Bernoulli offers competing arguments why the expected letter was never written (e.g. his brother might be dead), indicating a force of proof less than 1 for each of the competing arguments. 
\vspace{2pt}

In this context, Hailperin \cite{hailperin1996sentential} interprets Bernoulli as measuring the probabilities of A and $A \implies C$ to assess the probability of provability of C from A. For simplicity, Hailperin recommends joint probabilities and independence, but he must then assess separately probabilities of A and $A \implies C$, with different measures. The probability of A is 1. If A and C are to be confined to the same Boolean algebra, the probability of $A \implies C$ and the joint probability of A and $A \implies C$ are also 1. But there is no procedure in probability theory to span two Boolean algebras in such a calculation. Hailperin needs a better explanation at this point of analysis or this gap counts as an error.
\vspace{2pt}

The gap in Hailperin’s explanation is the starting point of our approach. We propose two new closely-related idioms of uncertain reasoning, uncertain inference arguments and uncertain implication arguments, that directly address the question of force of proof of arguments from propositions of one Boolean algebra to propositions of another. A sure sign of success in such an approach is to axiomatize consistency of such assignments in the style of Kolmogorov’s axioms for probability in an indefeasible way. We have taken this imaginative leap and the axioms must be challenged if our approach is to fail. We believe these axioms are “ineluctable” and don’t foresee such a challenge.

\section*{Notation and Definitions}
Let $\mathscr{A}$ be a finite Boolean algebra. $F_{\mathscr{A} }$ and $T_{ \mathscr{A} }$ represent the least and greatest element of $\mathscr{A}$. All Boolean algebras are finite in this report, though almost all results can be extended to non-finite cases.
\vspace*{4pt}

\noindent If A is an element of $\mathscr{A}$, let $A^{C}$ represents the complement of A, $n(A)$ represents the number of elements in A or the distance to the least element of $\mathscr{A}$. 
\vspace*{4pt}

\noindent If $A_1$ and $A_2$ are elements of $\mathscr{A}$, $A_1 \implies A_2$ means $A_1^{C} \vee A_2 = T_{\mathscr{A} }$.
\vspace{4pt}

\noindent If $A_1$ and $A_2$ are elements of $\mathscr{A}$, an \textbf{argument} from $A_1$ to $A_2$ is a triple, $(A_1 , A_2 , p(A_1^{C} \vee A_2)$, where p: $A \rightarrow [0, 1]$ is a
probability function. $p(A_1^{C} \vee A_2)$ is the \textbf{force of proof} of the argument.
\vspace*{4pt}

\noindent If $\mathscr{A}$ and $\mathscr{B}$ are Boolean algebras, an \textbf{argument} from $\mathscr{A}$ to $\mathscr{B}$ is a function:
\[FP: \mathscr{A} \times \mathscr{B} \rightarrow [0 ,1]\]
satisfying conditions: for all $A, A_1 , A_2$ in $\mathscr{A}$ and $B, B_1 , B_2$ in $\mathscr{B}$,

\renewcommand{\labelenumi}{\roman{enumi}.}
\begin{enumerate}
\item $FP(F_{\mathscr{A} } , B) = 1$
\item $FP(A, T_{\mathscr{B} } ) = 1$
\item $FP(T_{\mathscr{A} } , F_{\mathscr{B } } ) = 0$
\item If $A_1 \implies A_2$, $FP(A_1 , B) \geq FP(A_2 , B)$
\item If $B_1 \implies B_2$, $FP(A, B_1) \leq  FP(A, B_2 )$
\end{enumerate}

\noindent If FP is an argument from $\mathscr{A}$ to $\mathscr{B}$, define $\overrightarrow{FP}: \mathscr{A} \times \mathscr{B} \rightarrow [0,1]$ by: for all A in $\mathscr{A}$, B in $\mathscr{B}$, 
\[\overrightarrow{FP}(A, B)  = \sum_{A \implies \overline{A}}  (-1)^{n(A) - n(\overline{A})} FP(\overline{A}, B)\]

\noindent If FP is an argument from $\mathscr{A}$ to $\mathscr{B}$, define $\overleftarrow{FP}$: $\mathscr{A} \times \mathscr{B} \rightarrow [0,1]$ by: For all A in $\mathscr{A}$, B in $\mathscr{B}$,
\[\overleftarrow{FP}(A, B) = \sum_{\overline{B} \implies B} (-1)^{n(B) - n(\overline{B})} FP(A, \overline{B})\]

\noindent If $\overrightarrow{FP} \geq 0$, FP is an \textbf{uncertain implication argument}. If $\overleftarrow{FP} \geq 0$, FP is an \textbf{uncertain inference argument}.
\vspace*{2pt}

\noindent If FP is neither an uncertain inference argument nor an uncertain implication argument, FP is said to be a \textbf{superficial argument}.
\vspace*{2pt}

\noindent If the values of FP are 0 or 1, the argument is said to \textbf{discrete}. If FP is an uncertain implication argument or an uncertain inference argument, the values of $\overrightarrow{FP}$ or $\overleftarrow{FP}$ must be discrete as well.
\vspace{2pt}

\noindent We shall see that uncertain implication arguments can be composed, uncertain inference arguments can be composed, but superficial arguments cannot be composed. That is to say, a superficial argument is not entirely without merit: it can be advanced to ``make a point'', but cannot be used in a chain of reasoning.
\vspace*{4pt}

\begin{theorem}

In particular, from the axioms and definitions it is easily seen that:

\[\sum_{A \in \mathscr{A}} \overrightarrow{FP} (A, B) = 1\]

\[\textnormal{and}\]

\[\sum_{B \in \mathscr{B}} \overleftarrow{FP} (A, B) = 1\]

\end{theorem}

\noindent Thus, if FP is an uncertain implication or an uncertain implication argument then $\overrightarrow{FP}$ or $\overleftarrow{FP}$ is a (not-necessarily-normalized) mass function parametrized by elements of a Boolean algebra. A mass function is a map from a Boolean algebra to [0,1], the sum of whose values is 1.0, and is normalized if 0.0 is assigned to the minimal element (empty set/ false). Arguments can be viewed as mass functions parametrized by elements of Boolean algebras.

\vspace*{2pt}

\noindent If FP is an an argument from Boolean algebra $\mathscr{A}$ to Boolean algebra $\mathscr{B}$, and $FP(A, B) > 0$ implies $A = F_{\mathscr{A} }$, $A = T_{\mathscr{A} }$,  $B = F_{\mathscr{B} }$ or $B = T_{\mathscr{B} }$ for all A in $\mathscr{A}$ and B in $\mathscr{B}$, then FP is said to be \textbf{not probative}. Otherwise, FP is said to be \textbf{probative}.

\vspace*{2pt}

\noindent If FP is not probative, $\mathscr{A}$ and $\mathscr{B}$ are said to be \textbf{tangential with respect to FP}. Otherwise, $\mathscr{A}$ and $\mathscr{B}$ are said
to be \textbf{entangled with respect to FP}.

\vspace*{2pt}

\noindent Arguments are subjective constructions associated with voluntary deliberation, but most pairs of Boolean algebras would be treated as tangential if this deliberation were forced. E.g a Boolean algebra of propositions about contemporary music would almost certainly be treated as tangential when coupled with a Boolean algebra of propositions about metastable states in physics. However, a Boolean algebra of propositions about ethnicity and a Boolean algebra of propositions about nationality have a better chance of being treated as entangled by a social scientist.

\section*{Contrapositive Arguments}

If $\mathscr{A}$ and $\mathscr{B}$ are Boolean algebras, and FP is an argument from $\mathscr{A}$ to $\mathscr{B}$, then the contrapositive of FP,
\[C(FP): \mathscr{B} \times \mathscr{A} \rightarrow [0,1] \] 
is defined by: for A in $\mathscr{A}$, B in $\mathscr{B}$,

\[C(FP)(B, A) = FP(A^{C}, B^{C})\]
\vspace*{4pt}
\noindent The following are easily proven:
\begin{enumerate}
\item C(FP) is an argument from $\mathscr{B}$ to $\mathscr{A}$;
\item C(C(FP)) = FP;
\item if FP is an uncertain inference argument, C(FP) is an uncertain implication argument
\item if FP is an uncertain implication argument, C(FP) is an uncertain inference argument.
\end{enumerate}

\section*{The Prototypical Example}

Our constructions gain legitimacy if they are consistent with probability of provability in the case of a single Boolean algebra, and indeed they are:

\begin{theorem}
Let $\mathscr{A}$ be a Boolean algebra, $p_{\mathscr{A}}$ a probability measure on $\mathscr{A}$. Define $FP_{p_{\mathscr{A} } }: \mathscr{A} \times \mathscr{A} \rightarrow [0,1]$ by: for $A_1$ , $A_2$ in $\mathscr{A}$,
\[FP_{p_{\mathscr{A} } }(A_1 , A_2 ) = p_{\mathscr{A} } ( A_1^{C} \vee A_2)\]

\noindent Then $FP_{p_{\mathscr{A} } }$ is an uncertain implication argument and an uncertain inference argument.
\end{theorem}

An elementary but tedious proof of Theorem 2 is postponed to an appendix.

\section*{A Class of Uncertain Implication Arguments that are also Uncertain Inference Arguments:}

Let $\mathscr{A}$ be the Boolean algebra of subsets of set $\left\lbrace a_1, a_2, \dots , a_m \right\rbrace$ and $\mathscr{B}$ be the Boolean algebra of subsets of
$\left\lbrace b_1, b_2, \dots , b_n \right\rbrace$ for positive integers m and n. For $1 \leq i \leq m$ and $1 \leq j \leq n$, define $FP(\left\lbrace a_i \right\rbrace, \left\lbrace b_j \right\rbrace) = p_{i,j} \geq 0$ in such a way that $\sum_{j=1}^{j=n}  p_{i,j} = 1$, for $1 \leq i \leq m$. For $\left\lbrace a_i \right\rbrace$ a singleton in $\mathscr{A}$ and B in $\mathscr{B}$, let $FP(\left\lbrace a_i \right\rbrace, B ) = \sum_{ b_j \in B} p_{i,j}$. For A in $\mathscr{A}$ and B in $\mathscr{B}$, let $FP(A, B) = \Pi_{ a_i \in A} FP( \left\lbrace a_i \right\rbrace, B)$.
\vspace{4pt}

\noindent It follows easily that $\overrightarrow{FP} (A, B)= FP(A, B) * \Pi_{ \left\lbrace a_i \in A^{C} \right\rbrace }  1.0 - FP( \left\lbrace a_i  \right\rbrace, B)$ lies in [0,1]. (This
generalizes: 1 - x - y + xy = (1 - x)(1 - y) and x - xy = x(1 - y)). Thus FP is an uncertain implication argument . If A is $F_{\mathscr{A}}$, generally $\overleftarrow{FP} (A, B)$ is \textbf{not} 0, so the mass functions associated with elements of $\mathscr{A}$ are not necessarily normalized.
\vspace{4pt}

\noindent In computing $\overleftarrow{FP} (A, B)$ for A in $\mathscr{A}$ and B in $\mathscr{B}$, we must expand multivariate polynomials in variables $p_{i , j}$, as defined in the previous paragraph.
\begin{lemma}
For real numbers, $x_{1,1} , x_{2,1} , x_{2,1} , x_{2,2}$,
\begin{enumerate}
\item the monomials in the expansion of the polynomial $(x_{1,1} + x_{2,1} )(x_{2,1} + x_{2,2} )$ appear in the expansion of the polynomial $(x_{1,1} + x_{1,2} + x_{1,3} )(x_{2,1} + x_{2,2} + x_{2,3} )$ for any real numbers $x_{1,3}$, $x_{2,3}$.
\item any product of linear polynomials whose monomials contain the monomials of the expansion of polynomial $(x_{1,1} + x_{1,2} )(x_{2,1} + x_{2,2} )$ has the form $(x_{1,1} + x_{1,2} + x_{1,3} )(x_{2,1} + x_{2,2} + x_{2,3} )$ for some real numbers $x_{1,3} , x_{2,3}$.
\end{enumerate}
\end{lemma}

\noindent From an obvious generalization of the Lemma, It can be shown that for $B_1$ and $B_2$ disjoint, monomials in the expansion of $FP(A, B_1 )$ will appear in the expansion of $FP(A, B_1 \vee B_2 )$ and not in the expansion of $FP(A,B)$ for any B that cannot be represented as $B_1 \vee B_2$. In the expansion of $\overleftarrow{FP}(A, B)$, for $B_1 \vee B_2$ contained in B, these monomials will appear with alternating signs, with increasing numbers of instances as $B_2$ increases in size. The classic binomial expansion $(1 - 1)^{ n(B) - n(B_1)} = 0 = \sum_{ B_2 \subseteq  B - B_1}
\begin{pmatrix}
n(B - B_1) \\
n(B_2)
\end{pmatrix} (-1)^{ n(B-B_1) - n(B_2)}$ ensures that the monomials will appear in $\overleftarrow{FP}(A, B)$ with 0
coefficient, unless B and $B_1$ are equal, when the coefficients are 1. In any case, the sum of these monomials is the sum of some of the monomials of FP(A, B), and is therefore in [0, 1], since FP(A, B) is in [0,1]. Moreover, the monomials of FP(A, B) have coefficient 1 in $\overleftarrow{FP}(A, B)$, so they do not contribute to $\overleftarrow{FP}(A, F_B)$ and the mass function $\overleftarrow{FP}(A, \cdot)$ is normalized.

\section*{Categories of Mass Functions}

Let $\mathscr{A}$ be a Boolean algebra, $m_{\mathscr{A} } : \mathscr{A} \rightarrow [0,1]$ be a mass function. Let $\mathscr{B}$ be also a Boolean
algebra and FP(A, B) an uncertain inference argument. Let $m_{ \mathscr{B} } : \mathscr{B} \rightarrow [0, 1]$ be defined by:

\[m_{ \mathscr{B} } (B) = \sum_{A \in \mathscr{A}} m_{ \mathscr{A} }(A) * \overleftarrow{FP}(A, B)\]

\noindent It follows that $\sum_{ B \in \mathscr{B} } m_{ \mathscr{B} } (B) = \sum_{B \in \mathscr{B} } \sum_{A \in \mathscr{A}} m_{ \mathscr{A} }(A) * \overleftarrow{FP}(A, B) = 1$ and therefore
$m_{ \mathscr{B} } : \mathscr{B} \rightarrow [0,1]$ is a mass function.
\vspace*{2pt}

\noindent If $\overleftarrow{FP}(A,\cdot)$ is normalized, $m_{ \mathscr{B} } (F_{\mathscr{B} } ) = 0$, and $m_{ \mathscr{B} }$ is normalized, whether or not $m{ \mathscr{A} }$ is.

\vspace*{2pt}

\noindent On the other hand, let $\mathscr{B}$ be a Boolean algebra, $m_{ \mathscr{B}} : \mathscr{B} \rightarrow [0,1]$ be a mass function. Let $\mathscr{A}$ also be a Boolean algebra and FP(A, B) an uncertain implication argument. Let $m_{ \mathscr{A} } : \mathscr{A} \rightarrow [0, 1]$ be defined by:

\[m_{ \mathscr{A} }(A) = \sum_{B \in \mathscr{B}} m_{ \mathscr{B} } (B) * \overrightarrow{FP} (A, B)\]

\noindent It follows that $\sum_{A \in \mathscr{A} } m_{ \mathscr{A} }(A) = \sum_{A \in \mathscr{A} } \sum_{B \in \mathscr{B}} m_{ \mathscr{B} } (B) * \overrightarrow{FP} (A, B) = 1$, and therefore that $m_{ \mathscr{A} } (A): \mathscr{A} \rightarrow [0,1]$ is a mass function.
\vspace*{2pt}

\noindent If $\overrightarrow{FP} (\cdot,B)$ is normalized, $m_{ \mathscr{A} } (F_{\mathscr{A}} ) = 0$, and $m_{ \mathscr{A} }$ is normalized, whether or not $m_{ \mathscr{B} }$ is.
\vspace*{2pt}

\noindent Suppose $\mathscr{A}, \mathscr{B},$ and $\mathscr{C}$ are Boolean algebras, $FP_{ \mathscr{A} , \mathscr{B} } : \mathscr{A} \times \mathscr{B} \rightarrow [0, 1]$ and $FP_{ \mathscr{B} , \mathscr{C}} : \mathscr{B} \times \mathscr{C} \rightarrow [0, 1]$ are uncertain inference arguments. Let $m_{ \mathscr{ A} } : \mathscr{A} \rightarrow [0,1]$ be a mass function. Let $m_{ \mathscr{C} } : \mathscr{C} \rightarrow [0,1]$ be the mass function defined by:

\[m_{ \mathscr{C} } (C) = \sum_{B \in \mathscr{B}} \sum_{A \in \mathscr{A}} m_{ \mathscr{A} } (A) * \overleftarrow{FP}_{ \mathscr{A}, \mathscr{B} } (A,B) * \overleftarrow{FP}_{ \mathscr{B} , \mathscr{C}} (B,C)\] 

\noindent for C in $\mathscr{C}$.

This construction defines the composition of $\overleftarrow{FP}_{ \mathscr{B}, \mathscr{C} } (B, \cdot)$ with $\overleftarrow{FP}_{ \mathscr{A}, \mathscr{B} } (A, \cdot)$
\vspace*{2pt}

\noindent Suppose $\mathscr{A}, \mathscr{B}$, and $\mathscr{C}$ are Boolean algebras, $FP_{ \mathscr{A} , \mathscr{B} } : \mathscr{A} \times \mathscr{B} \rightarrow [0, 1]$ and $FP_{ \mathscr{B} , \mathscr{C}} : \mathscr{B} \times \mathscr{C} \rightarrow [0, 1]$ are uncertain
implication arguments. $m_{ \mathscr{A} } : \mathscr{A} \rightarrow [0,1]$ be the
mass function defined by:

\[m_{ \mathscr{A} } (A) = \sum_{B \in \mathscr{B}} \sum_{C \in \mathscr{C} } m_{ \mathscr{C} } (C) * \overrightarrow{FP}_{ \mathscr{A}, \mathscr{B} } (A,B) * \overrightarrow{FP}_{ \mathscr{B}, \mathscr{C} } (B,C) \] 
for A in $\mathscr{A}$.

\vspace*{2pt}

\noindent This construction defines the composition of $\overrightarrow{FP}_{ \mathscr{B}, \mathscr{C} } (\cdot,C)$ with $\overrightarrow{FP}_{ \mathscr{A}, \mathscr{B} } (\cdot,B)$.
\vspace*{4pt}

\noindent \textbf{Morphisms in a category must compose}. We have therefore constructed two categories of mass functions, associated with implication arguments and inference arguments and two sub-categories of normalized mass functions, associated with normalized implication arguments and normalized inference arguments.
\vspace*{2pt}

\noindent It is noteworthy that a category of normalized mass functions is equivalent to a category of belief functions, in the sense of Shafer \cite{shafer1976mathematical}, which does not use the formalism of category theory. Elsewhere (\cite{shafer1990perspectives}) Shafer introduces compatibility relations between Boolean algebras specifically for the purpose of transferring weight from functions on the first Boolean algebra to belief functions of the second. These can be interpreted as morphisms in the sense introduced here, and will be discussed in the next section.
\vspace*{2pt}

\noindent In (\cite{shafer1990perspectives}) Shafer indicates a revised view (relative to \cite{shafer1976mathematical}) that these morphisms are the primary subject of his theory of uncertain reasoning, not the belief functions apart from their association with compatibility relations. Therefore, it can be said that our implementation of Bernoulli’s project encompasses a strict generalization of the latest official version of Dempster-Shafer theory. It remains to be seen in the next section what is the scope of this generalization.

\section*{Compatibility Relations as Morphisms}

Let $\mathscr{A}$ and $\mathscr{B}$ be Boolean algebras. A compatibility relation between $\mathscr{A}$ and $\mathscr{B}$ is a relation CR in $\mathscr{A} \times \mathscr{B}$
satisfying:

\begin{enumerate}
\item $F_{ \mathscr{A} }$ CR $B$ for B in $\mathscr{B}$ if and only if $B = F_{ \mathscr{B} }$
\item $T_{ \mathscr{A} }$ CR $B$ for B in $\mathscr{B}$ if B is not $F_{\mathscr{B}}$
\item $A$ CR $F_{ \mathscr{B} }$ for A in $\mathscr{A}$ if and only if $A = F_{ \mathscr{A} }$
\item $A$ CR $T_{ \mathscr{B} }$ for A in $\mathscr{A}$ if A is not $F_{ \mathscr{A}}$
\item $A$ CR $B_1 \vee B_2$ if and only if $A$ CR $B_1$ or $A$ CR $B_2$ for A in $\mathscr{A}$ and $B_1, B_2$ in $\mathscr{B}$
\item $A$ CR $B_1 \wedge B_2$ if and only if $A$ CR $B_1$ and $A$ CR $B_2$ for A in $ \mathscr{A}$ and $B_1 , B_2$ in $\mathscr{B}$
\item $A_1 \vee A_2$ CR $B$ if and only if $A_1$ CR $B$ or $A_2$ CR $B$ for $A_1 , A_2$ in $\mathscr{A}$ and B in $\mathscr{B}$
\item $A_1 \wedge A_2$ CR $B$ if and only if $A_1$ CR $B$ and $A_2$ CR $B$ for $A_1 , A_2$ in $\mathscr{A}$ and B in $\mathscr{B}$
\end{enumerate}

If CR is a compatibility relation between $\mathscr{A}$ and $\mathscr{B}$, define $FP_{CR} : \mathscr{A} \times \mathscr{B} \rightarrow [0, 1]$ as follows:
\vspace*{2pt}

\noindent For A in $\mathscr{A}$, B in $\mathscr{B}$,
\begin{enumerate}
\item $FP_{CR} (A, B) = 1$ if $A$ CR $B$ and not( $A$ CR $B^{C}$)
\item $FP_{CR} (A, B) = 1$ if $A = F_{ \mathscr{A} }$
\item $FP_{CR} (A, B) = 0$ if not i or ii.
\end{enumerate}
It should be noted that in the case $\mathscr{A} = \mathscr{B}$, $FP_{CR} (A, B) = 1$ if and only if $A^{C} \wedge B = T_{ \mathscr{A} }$. In the general case, $FP_{CR}$ is easily seen to satisfy the axioms of an uncertain implication argument. 
\vspace*{2pt}

\noindent These compatibility relations illustrate discrete arguments. Since the computation of $\overrightarrow{FP}_{CR} (A, B)$ from $FP_{CR} (A, B)$ involves only sums of $0$’s and $1$’s, and values lie between $0$ and $1$, all values of $\overrightarrow{FP}_{CR} (A, B)$ must
be $0$ or $1$.

\vspace*{2pt}

\noindent In Shafer’s most recent summary (\cite{shafer2008non}) of his views on Dempster-Shafer theory, he claims the principal focus should be such compatibility relations between Boolean algebras decorated with probability measures (for $ \mathscr{A}$) and belief functions (for $\mathscr{B}$).
\vspace*{2pt}

\noindent This sparse set of ``morphisms'' sits inside our category of normalized mass functions with normalized uncertain implication arguments, almost all of whose innumerable morphisms were
introduced here for the first time. It remains to be seen if this greatly expanded scope will be helpful in applications, but it represents a theoretical breakthrough in belief function calculus.

\section*{Summary}
We have exhumed informal notes from Jakob Bernoulli’s famous tract, \textbf{Ars Conjectandi}. We have exploited them, but not for the purpose of explaining $17^{th}$ century styles of reasoning about uncertainty, as others have. Rather we have exploited them to develop new idioms of uncertain reasoning closely related to and substantially generalizing one of the most prominent contemporary explanatory models, Shafer’s evidence theory, with the prospect of greatly expanding its scope for applications. As a critical step, this report offers the first explicit solution to a central problem of uncertain reasoning posed by Bernoulli that has been unsolved for 300 years, in spite of its publication in a classic text, well-known to the mathematical community during those years.

\subsection*{Appendix: Proof of Theorem 2}
\begin{NoNumTheorem}
Let $\mathscr{A}$ be a Boolean algebra, $p_{\mathscr{A} }$ a probability measure on $\mathscr{A}$. Define $FP_{p_{\mathscr{A} } }: \mathscr{A} \times \mathscr{A} \rightarrow [0,1]$ by: for $A_1$, $A_2$ in $\mathscr{A}$, 

\[FP_{ p_{\mathscr{A} } } (A_1, A_2) = p_{\mathscr{A}} ( A_2 \vee A_1^{C}).\]

\noindent Then $FP_{ p_{\mathscr{A} } }$ is an uncertain implication argument and an uncertain inference argument.

\end{NoNumTheorem}

\begin{lemma}
\leavevmode
\begin{enumerate}
\item If n is a positive integer, $\displaystyle{\sum_{m = 0}^{n} \begin{pmatrix}
n \\
m
\end{pmatrix} (-1)^{n - m} = 0}$
\item If $\mathscr{A}$ is a Boolean algebra, $p: \mathscr{A} \rightarrow [0,1]$ is a probability measure and A is an element of $\mathscr{A}$, then
\[\sum_{ \overline{A} \subseteq A} (-1)^{n(A) - n(\overline{A}) } p(\overline{A}) \geq 0.0\]
\end{enumerate}
\end{lemma}
 
(These are well-known combinatorial identities).

\subsubsection*{Uncertain Implication}

From the definitions: 
\[\overleftarrow{FP_{ p_{\mathscr{A} } } }(A_1 , A_2) = \sum_{ A_1 \subseteq A_3} (-1)^{ n(A_1) - n(A_3) } FP(A_3 , A_2 ) =
\sum_{A_1 \subseteq A_3 } (-1)^{n(A_1 ) - n(A_3)} p_{\mathscr{A}} (A_2 \vee A_3^{C} )\]

\noindent If $A_1 = T_{\mathscr{A}}$ then this sum reduces down to

\[\sum_{T_{\mathscr{A}} \subseteq A_3 } (-1)^{n(T_{\mathscr{A}} ) - n(A_3)} p_{\mathscr{A}} (A_2 \vee (A_3)^{C} ) = p_{\mathscr{A}} (A_2) \geq 0\]
Otherwise, letting $C_1 = A_1^{C}$, $C_3 = A_3^{C}$, this is

\begin{align*}
\sum_{C_3 \subseteq C_1}  (-1)^{n(C_1 ) - n(C_3 )} p_{\mathscr{A} } (A_2 \vee C_3 ) & =
\sum_{C_3 \subseteq C_1} (-1)^{n(C_1 ) - n(C_3 )} p_{\mathscr{A} } (A_2 ) +
\sum_{C_3 \subseteq C_1} (-1)^{n(C_1 ) - n(C_3 )} p_{\mathscr{A}} (C_3 \wedge A_2^{C}) \\
& = 0.0 + \sum_{C_3 \subseteq C_1} (-1)^{n(C_1 ) - n(C_3 )} p_{\mathscr{A}} (C_3 \wedge A_2^{C})
\end{align*}

\noindent For $C_3$ contained in $C_1$, consider subsets of $C_1$ related to $C_3$ as follows:
\[ \left\lbrace  C \subseteq C_1 | \;  C \wedge A_2^{C} = C_3 \wedge A_2^{C} \right\rbrace \] 

\noindent If $D = C_3 \wedge A_2^{C}$ these subsets of $C_1$ are described as 

\[\left\lbrace D \vee E | \; E \subseteq C_1 \wedge A_2 \right\rbrace\]

\noindent The last sum can therefore be redescribed as:

\[\sum_{D \subseteq C_1 \wedge A_2^{C}}  (-1)^{n(C_1 \wedge A_2^{C} ) - n(D)}
\sum_{E \subseteq C_1 \wedge A_2} (-1)^{n(C_1 \wedge A_2) - n(E)} p_{\mathscr{A} } (D)\] 

\noindent But the first factor is 0.0 (from the Lemma) unless $C_1 \wedge A_2$ is $F_{\mathscr{A} }$, when 

\[\sum_{D \subseteq C_2 \wedge A_2^{C}}  (-1)^{n(C_1 \wedge A_2 ) - n(D)}
\sum_{E \subseteq F_{\mathscr{A}} } (-1)^{n(F_\mathscr{A}) - n(E)} p_{\mathscr{A} } (D) = \sum_{D \subseteq A_2 \wedge A_1}  (-1)^{n(A_2 \wedge A_1 ) - n(D)}
(1) p_{\mathscr{A} } (D) \geq 0.0\] 
by a previous lemma. In any case, the second factor is $\geq 0$.

\subsubsection*{Uncertain Inference}

From the definitions: 
\[\overrightarrow{FP_{ p_{\mathscr{A} } } } (A_1 , A_2) = \sum_{ A_3 \subseteq A_2} (-1)^{ n(A_2) - n(A_3) } FP(A_1 , A_3 ) =
\sum_{A_3 \subseteq A_2 } (-1)^{n(A_2 ) - n(A_3)} p_{\mathscr{A}} (A_3 \vee A_1^{C} )\]
If $A_2 = F_{\mathscr{A}}$ then this sum reduces to
\[\sum_{A_3 \subseteq F_{\mathscr{A}}} (-1)^{n(F_{\mathscr{A}}) - n(A_3)} p_{\mathscr{A}} (A_3 \vee A_1^{C} ) = p_{\mathscr{A}} (A_1^{C}) \geq 0\]

\noindent Otherwise,
\begin{align*}
\sum_{A_3 \subseteq A_2}  (-1)^{n(A_2 ) - n(A_3 )} p_{\mathscr{A} } (A_3 \vee A_1^{C} ) & =
\sum_{A_3 \subseteq A_2} (-1)^{n(A_2 ) - n(A_3 )} p_{\mathscr{A} } (A_1^{C}) +
\sum_{A_3 \subseteq A_2} (-1)^{n(A_2 ) - n(A_3 )} p_{\mathscr{A}} (A_3 \wedge A_1) \\
& = 0.0 + \sum_{A_3 \subseteq A_2} (-1)^{n(A_2 ) - n(A_3 )} p_{\mathscr{A}} (A_3 \wedge A_1)
\end{align*}

\noindent For $A_3$ contained in $A_2$, consider subsets of $A_2$ related to $A_3$ as follows:

\[ \left\lbrace  A \subseteq A_2 | \;  A_3 \wedge A_1 = A \wedge A_1 \right\rbrace \] 

\noindent If $D = A_3 \wedge A_1$ these subsets of $A_2$ are described as:

\[\left\lbrace D \vee E | \; E \subseteq A_2 \wedge A_1^{C} \right\rbrace\]

\noindent The last sum can therefore be redescribed as:

\[\sum_{D \subseteq A_2 \wedge A_1}  (-1)^{n(A_2 \wedge A_1 ) - n(D)}
\sum_{E \subseteq A_2 \wedge A_1^{C}} (-1)^{n(A_2 \wedge A_1^{C}) - n(E)} p_{\mathscr{A}} (D)\] 

\noindent But the first factor is 0.0 (from the Lemma) unless $A_2 \wedge A_1^{C}$ is $F_{\mathscr{A} }$, when 
\[\sum_{D \subseteq A_2 \wedge A_1}  (-1)^{n(A_2 \wedge A_1 ) - n(D)}
\sum_{E \subseteq F_{\mathscr{A}}} (-1)^{n(F_{\mathscr{A}}) - n(E)} p_\mathscr{A} (D) = \sum_{D \subseteq A_2 \wedge A_1}  (-1)^{n(A_2 \wedge A_1 ) - n(D)}
(1) p_{\mathscr{A} } (D) \geq 0.0\] 
by a previous lemma. In any case, the second factor is $\geq 0$.

\pagebreak
\bibliographystyle{plain}
%
\end{document}